\documentclass[10pt,twocolumn,letterpaper]{article}

\usepackage{cvpr}
\usepackage{times}
\usepackage{epsfig}
\usepackage{graphicx}
\usepackage{amsmath}
\usepackage{amssymb}
\usepackage{subfigure}

\usepackage{algorithm}
\usepackage{amsmath, amssymb, amsthm, caption,bm}
\usepackage{algpseudocode}
\usepackage{amsmath}
\usepackage{multirow}
\usepackage{subfigure}

\usepackage[breaklinks=true,bookmarks=false]{hyperref}
\cvprfinalcopy 
\def\cvprPaperID{****} 

\begin{document}

\title{AI Challenger : A Large-scale Dataset for Going Deeper in Image Understanding}
\author{Jiahong Wu$^{\dagger1}$, He Zheng$^{\dagger2}$, Bo Zhao$^{\dagger3}$, Yixin Li$^{\dagger3}$, Baoming Yan$^{\dagger3}$, Rui Liang$^{\dagger1}$\\
Wenjia Wang$^{3}$, Shipei Zhou$^{1}$, Guosen Lin$^{3}$, Yanwei Fu$^{4}$, Yizhou Wang$^{3}$, Yonggang Wang$^{\ddagger1}$\\
\normalsize \small$^1$Sinovation Ventures, $^2$University of Chinese Academy of Sciences,\\
\normalsize \small$^3$Peking University, $^4$School of Data Science, Fudan University\\
\small$\dagger$ Equal contribution. $\ddagger$  Corresponding author. This work is mainly done in Sinovation Ventures.\\
{\small\{wujiahong, liangrui, wangyonggang\}@chuangxin.com, zhenghe15@csu.ac.cn, shipeiz@andrew.cmu.edu}\\
{\small\{bozhao, liyixin, bmyan, wenjiawang, linguosen, Yizhou.Wang\}@pku.edu.cn, yanweifu@fudan.edu.cn}
}
\maketitle

\begin{abstract}
Significant progress has been achieved in Computer Vision by leveraging large-scale image datasets. However, large-scale datasets for complex Computer Vision tasks beyond classification are still limited. This paper proposed a large-scale dataset named AIC (AI Challenger) with three sub-datasets, human keypoint detection (HKD), large-scale attribute dataset (LAD) and image Chinese captioning (ICC). In this dataset, we annotate class labels (LAD), keypoint coordinate (HKD), bounding box (HKD and LAD), attribute (LAD) and caption (ICC). These rich annotations bridge the semantic gap between low-level images and high-level concepts. The proposed dataset is an effective benchmark to evaluate and improve different computational methods. In addition, for related tasks, others can also use our dataset as a new resource to pre-train their models.
\end{abstract}


\section{Introduction}
The recent progress achieved in Computer Vision tasks largely rely on deep neural networks\cite{2016resnet,szegedy2015going} and big data, such as ImageNet\cite{NIPS2012_4824}, MSCOCO\cite{2014MSCOCO}, Scene Understanding (SUN)\cite{xiao2010sun} and Flickr30k\cite{2014Flickr30k}. Most existing datasets focus on traditional (object or scene) classification and recognition tasks. Many images are annotated with only labels and bounding boxes. However predicting labels and bounding boxes of objects are far away from deep understanding of images. Those datasets with rich annotations, such as human keypoints, attributes and captions, are a small fraction of existing datasets and have a small scale. In human keypoint detection task, MSCOCO\cite{2014MSCOCO} and MPII\cite{mpii} provided no more than 200k labelled images. The sum of images in frequently used attribute datasets (CUB, SUN Attribute, aP/aY and AwA ) is only 72k. Currently, Flickr8k-cn\cite{2016Flickr8k_cn} provides 8k Chinese captions of images, however, because annotation does not specify any rules on wording, some captions may not contain all salient objects and some may not express the relationship between objects, which will result in a lack of information for training or evaluating methods.

\begin{figure}[t]
\begin{center}
   \includegraphics[width=1\linewidth]{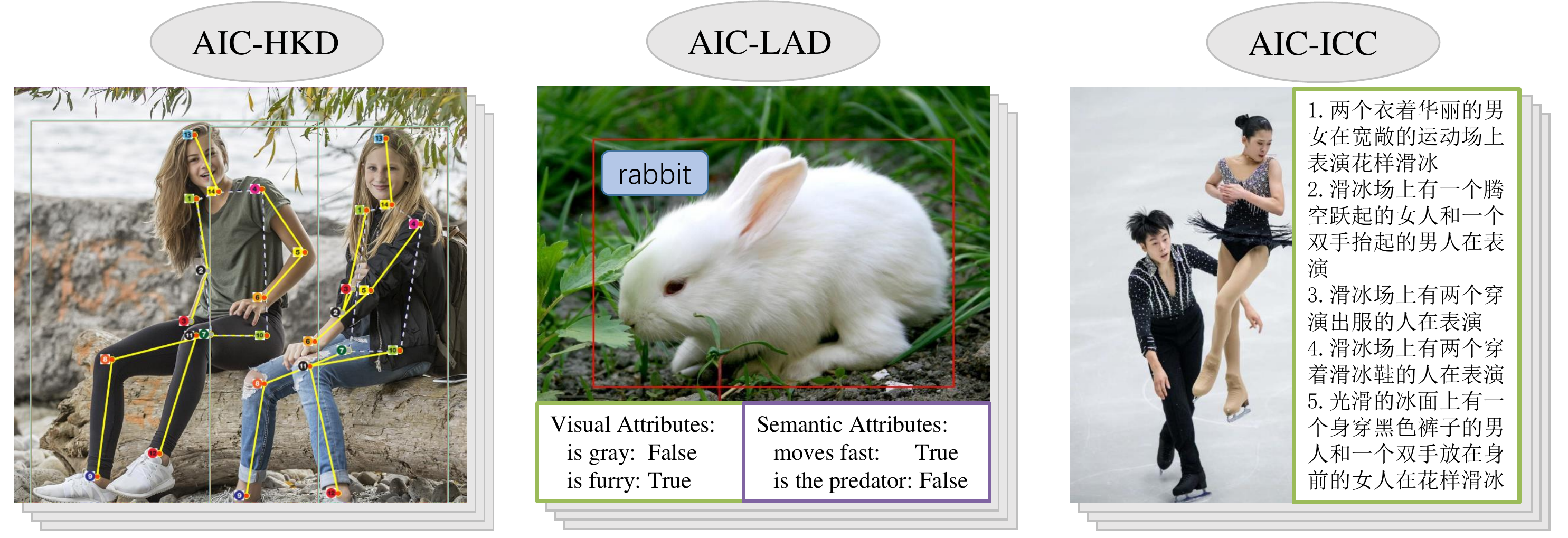}
\end{center}
   \caption{\small{Examples of the proposed datasets.}}
\label{example_dataset}
\end{figure}

The goal of this paper is to go deeper in image understanding by providing a dataset for three more comprehensive tasks, namely, human keypoint detection, attribute based zero-shot recognition and image Chinese captions (see Fig.\ref{example_dataset}). These three tasks focus on the concept of daily life for ordinary people. In human keypoint detection task, we try to annotate and predict the keypoints of people, which is a fundamental task for capturing and understanding human activities. In attribute based zero-shot recognition task, we are inspired by human being's learning ability, that people can learn new concepts from descriptions, and we annotate attributes of objects for implementing zero-shot recognition. In image Chinese captioning task, we try to understand the relation between objects in the image by captioning and we annotate Chinese captions for scenes of people's daily life.

To build such dataset, we first design the scene and object categories. Then the raw images are crawled from the Internet by querying label names in search engine. Then, these images are divided into three classes, namely "useless", "single-object", and "multi-object". We use only single-object images to build the attribute dataset for zero-shot recognition, while we use both single-object images and multi-object images for keypoint detection and Chinese captioning. The whole dataset contains 300,000 images (annotated with key points for main characters) for keypoint detection, 81,658 images (annotated with labels, bounding boxes, and attributes (partially)) for zero-shot recognition and 300,000 images ,annotated with 5 Chinese captions per image, for Chinese captioning. We should emphasize that there are more than 95\% overlap between keypoint images and captioning images. Hence, this is a good resource to investigate how to jointly deal with two different visual tasks.

There are three main contributions in this paper. 1) Our dataset provides a new benchmark to evaluate methods in the three tasks. 2) The huge dataset is a new resource for pre-training models. 3) To our best knowledge, this is the first large-scale image Chinese captioning dataset.


\section{Human Skeletal System Keypoint Detection}

\subsection{Overview}

Human Skeletal System Keypoint Detection plays an important role in several computer vision tasks, such as pose estimation, activity recognition and abnormal action detection. Unfortunately, due to the unknown number, position and scale of human figure in the image, along with the interactions and occlusions that may occur between people, human keypoint detection can be a real challenging task.

Recent human keypoint detection approaches can be roughly divided into two categories\cite{openpose}: top-down\cite{Iqbal2016Multi, Gkioxari2014Using, Pishchulin2012Articulated, Sun2011Articulated} and bottom-up\cite{openpose, Bulat2016Human, DBLP:journals/corr/ChenY14c, Tompson2014Joint}. The main idea of a top-down scenario is to divide and conquer, which treats the multi-person keypoint detection problem as a human detection followed by a single person keypoint detection. On the other hand, a bottom-up method directly extracts human keypoints from the image and clusters the results into different humans.

In the last few years, the deep neural networks especially the Convolutional Neural Networks(CNN), have been widely used to detect and localize the human keypoints\cite{Newell2016Stacked, Ouyang2014Multi, Toshev2014DeepPose, Wei2016Convolutional}. To avoid over-fitting, such approaches require massive labelled data to train the deep neural networks. While existing datasets with human keypoint annotation like MSCOCO\cite{2014MSCOCO} and MPII\cite{mpii} provide only no more than two hundred thousand labelled images, here we introduce the Human skeletal system Keypoint Detection Dataset(HKD) which contains 300,000 high resolution images with multiple persons and various poses, and each person is labeled with a bounding box and 14 human skeletal keypoints. The comparison between datasets is shown in Tab.\ref{keypoint_datasets}.

\begin{table}[hbp]\small
\centering
\begin{tabular}{cccc}
\hline
Datasets & Images & Humans & Keypoints \\
\hline
MSCOCO\cite{2014MSCOCO} & 200k & 250k & 17 \\
MPII\cite{mpii} & 25k & 40k & 13 \\
HKD(Ours) & 300k & 700k & 14 \\
\hline
\end{tabular}
\caption{\small{The comparison of human keypoint datasets.}}\label{keypoint_datasets}
\end{table}

The rest of this section is organized as follows: we first describe how we collected and annotated the images, some dataset statistics are shown in the next subsection, then the evaluation metrics we designed for the task is described, finally we introduce the baseline model and conduct some experiments.

\subsection{Data Annotation}

The annotation pipeline for HKD data set can be divided into three major parts, which are image filtering, human bounding box labeling and human skeletal keypoints labeling.

Similar to the SCH and the ICC dataset, images in the HKD dataset are collected from Internet search engines. So the first step is to remove inappropriate images out of the HKD dataset. These may include but are not limited to those images containing famous politicians, domestic police forces, sexual contents, violence or other inappropriate actions. In addition, we eliminate images where all human figures are too small(e.g. football players on the field taken from the top of stadium stand), or the ones that contain too many human figures (e.g. the crowd on the stadium stand) from our dataset.

The next step is to label human figures with bounding boxes. The bounding box should stay as close to the subject as possible, and in the mean time, contain all visible parts of this human figure. Note that not all humans in images are labelled with a bounding box. We skip the small human figure whose body parts are hard to distinguish, and the vague ones whose body contours are hard to recognize, because we want the algorithm to focus on detecting the most significant human body instead of all the humans in the image.

The final and the most important step is to label the locations and types of human skeletal keypoints for each human with a bounding box from the previous annotation stage. For each human, we labeled 14 human skeletal keypoints, and the numeric order of these keypoints is : 1-right shoulder, 2-right elbow, 3-right wrist, 4-left shoulder, 5-left elbow, 6-left wrist, 7-right hip, 8-right knee, 9-right ankle, 10-left hip, 11-left knee, 12-left ankle, 13-top of the head, and 14-neck. Each keypoint has one of three visibility flags: labeled and visible, labeled but not visible, or not labeled.

\subsection{Data Statistics}
\begin{figure}[htp]
\begin{center}
   \includegraphics[width=1\linewidth]{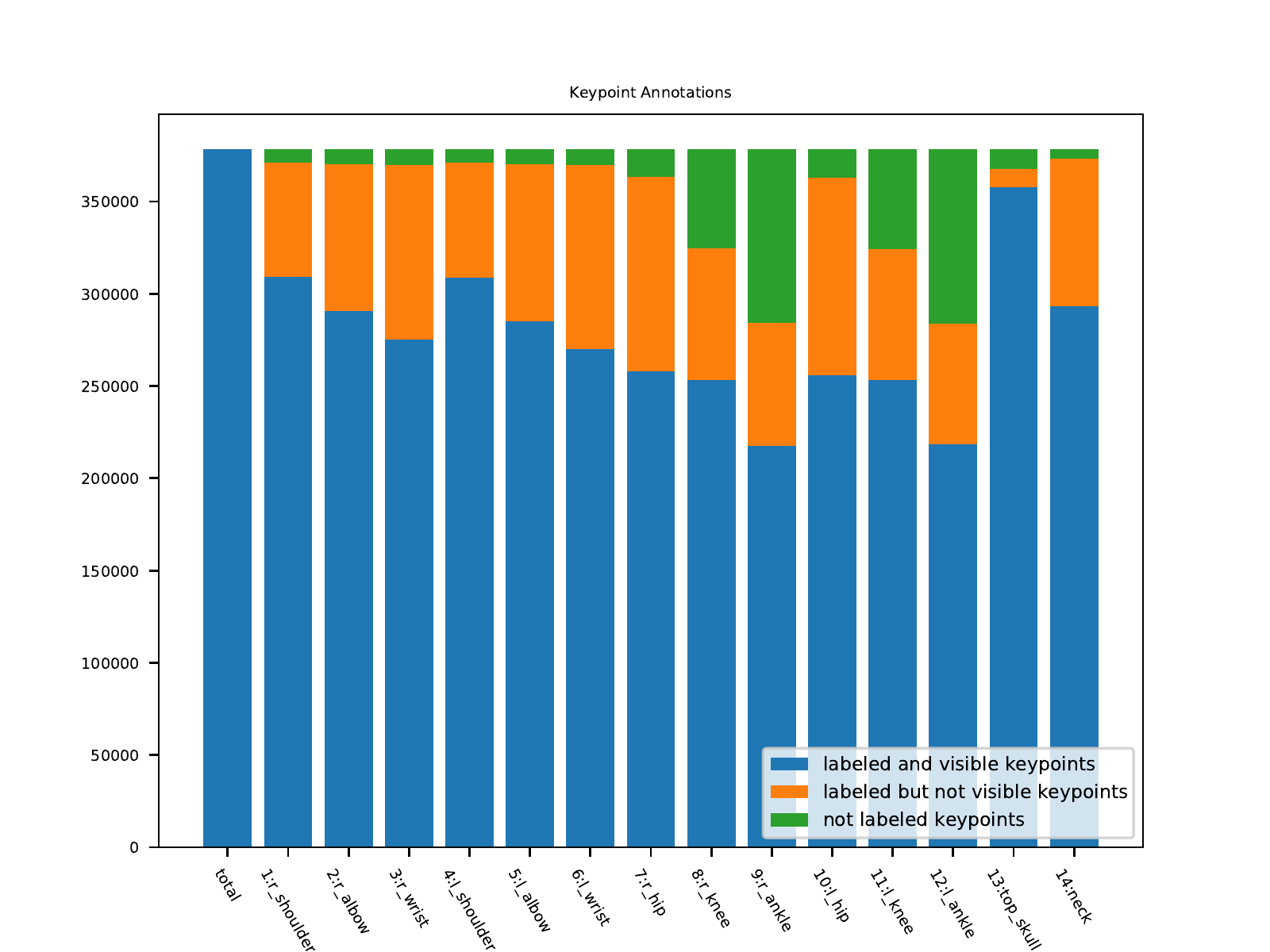}
\end{center}
   \caption{\small{The distribution of different type of keypoints.}}
\label{keypoint_fig2}
\end{figure}
We split the HKD dataset into training, validation, test A and test B with 70\%, 10\%, 10\% and 10\% ratio, which contain 210\,000, 30\,000, 30\,000 and 30\,000 images respectively. We only provide statistics on 210\,000 training data.

For the 210\,000 images in training set, there are 378\,374 human figures with almost 5 million keypoints. Among all the human keypoints we have labeled, 78.4\% of them are labeled as visible($v=1$) and the rest of them are labeled as not visible($v=2$). The distribution of different types of keypoints are shown in Fig.\ref{keypoint_fig2}

\begin{figure}[htp]
\begin{center}
   \includegraphics[width=1\linewidth]{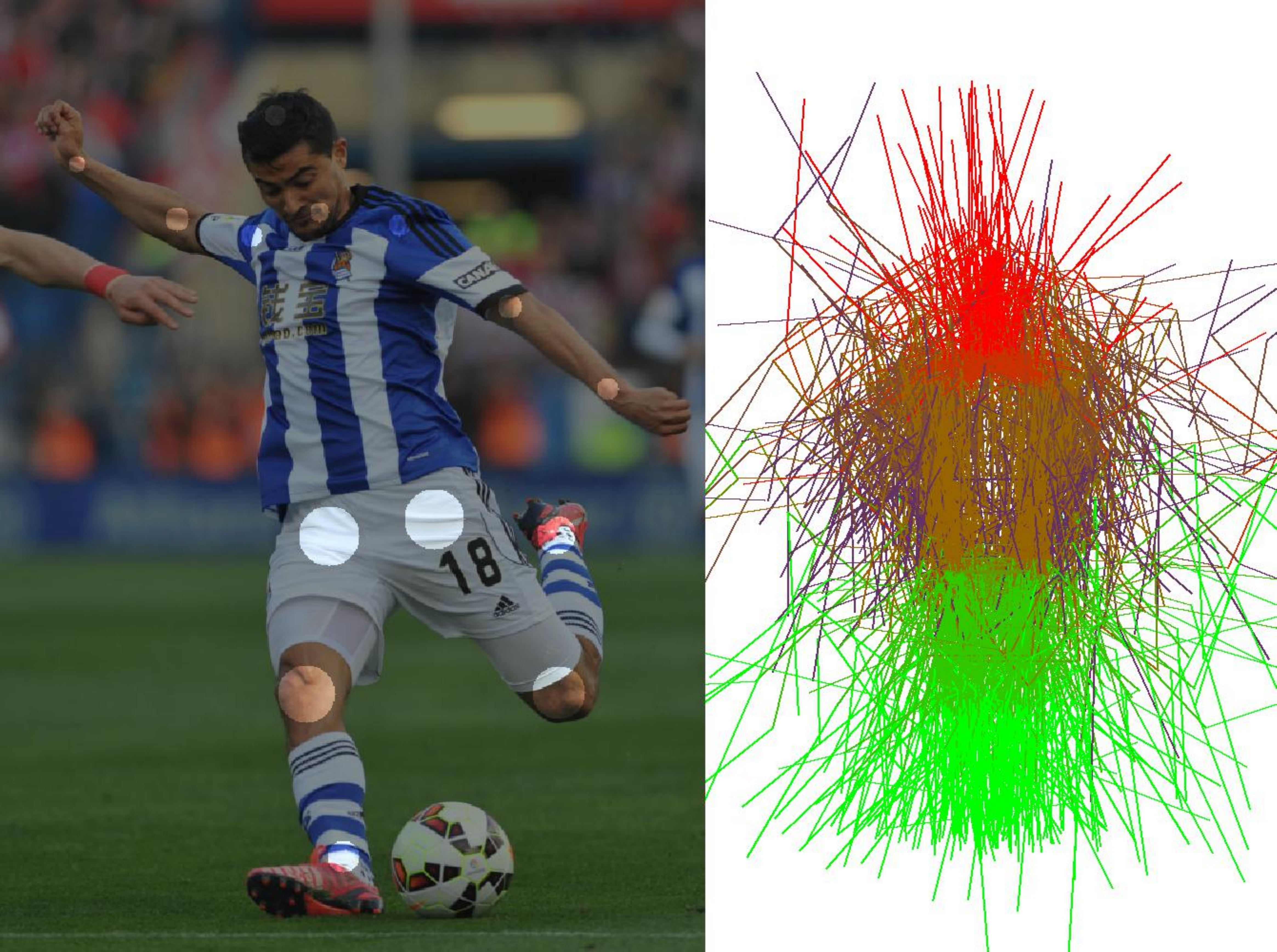}
\end{center}
   \caption{\small{Human Label Deviation and Pose Diversity. (a)The radius of bright circle is the human label deviation of corresponding keypoint type, which represents the difficulty of prediction. (b)To demonstrate the diversity of human poses in the HKD dataset, 100 human are randomly chosen and Human limbs are drawn after aligned.}}
\label{keypoint_fig1}
\end{figure}

Inconsistency in human-annotated keypoint locations is inevitable. We had 33 people labeled a same batch of 100 images to test the noise introduced by humans. In specific, we calculate the second central moment, which is the maximum likelihood estimation on standard deviation of the Euclidean distance between each type of keypoints and its center. The human label deviation is shown in Fig.\ref{keypoint_fig1}(a), where the radius of bright circle is the human label deviation of corresponding keypoint type. We can see that the upper body is labeled more accurately and the hips are generally more difficult to annotate. These human label deviation of different types of keypoints are used in evaluation metrics to measure the prediction difficulty, which will be introduced in the next subsection.

To demonstrate the diversity of human poses in the HKD dataset, 100 human annotations are randomly chosen from the training set. We apply the keypoint alignment by linear transformation, where the parameters of the transformation are set to make these 5 keypoints, right shoulder(1), left shoulder(4), right hip(7), left hip(10) and neck(14), have the same first moment(center) and second central moment(standard deviation). As shown in Fig.\ref{keypoint_fig1}(b), the most common poses are standing and sitting, while there are also quite a few other poses.

\subsection{Evaluation Metrics}

The evaluation metric of the human skeleton keypoint detection is similar to common object detection task, where the submission is scored in mean Average Precision (mAP). In common object detection tasks, Intersection over Union (IoU) is used to evaluate the similarity between a predicted bounding box and a ground truth bounding box. While in the human skeletal system keypoints detection task, we use Object Keypoint Similarity (OKS) proposed in \cite{2014MSCOCO} instead of IoU, to measure the similarity between the predicted keypoints and the ground truth keypoints.

The mAP score is the mean value of the Average Precision (AP) score under different OKS thresholds(0.50:0.05:0.95). The AP (Average Precision) score is calculated in the same way as in common object detection, but instead of IoU, OKS is used as similarity metric. Given the OKS threshold s, the AP under $s$ (AP@s) of the test results is predicted by the participants over the entire test set.

The OKS score is similar to the IoU score in common object detection task, which measures the similarity between the prediction and the ground truth. The main idea of OKS is the weighted Euclidean distance of the predicted keypoints and the ground truth keypoints, and for each human figure p, the OKS score is defined as follows:

$$OKS_p=\frac{\sum_{i}exp\left\{-d_{pi}^2/2s_p^2\sigma_{i}^2\right\}\delta\left(v_{pi}=1\right)}{\sum_{i}\delta\left(v_{pi}=1\right)}$$

Where $p$ is the index of human annotations; $i$ is the id number of the given human skeleton keypoint; $d_{pi}$ is the Euclidean distance between the predicted keypoint position and the ground truth; $s_p$ is the scale factor of human figure $p$, which is defined as the square root of the human bounding box area of human figure $p$; $\sigma_{i}$ is the normalized factor of the human skeletal keypoint, which is calculated by the standard deviation of human annotation result; $v_{pi}$ is the the visibility flag of the $i$ keypoint of the human figure $p$; $\delta\left(\cdot\right)$ is the Kronecker function, which means only visible human skeletal keypoints($v=1$)are considered during evaluation.

An evaluation script will be comming soon to facilitate offline evaluation.

\subsection{Baseline Model and Experiments}

We provide a basic approach to detect human skeletal keypoints in natural images as the baseline model of the HKD dataset. The most straightforward way is to adopt a top-down type method, that we first detect the humans in the image and then apply a single person keypoint detection method. The baseline model consists of three major parts: a human detector, a keypoint detector and a post-processing procedure to complete the task.

For detector we choose the Single Shot multibox Detector(SSD)\cite{ssd} pre-trained on Pascal VOC\cite{pascal_voc} and Mask R-CNN\cite{mask_rcnn} pre-trained on MSCOCO\cite{2014MSCOCO}. Since person is one of the defined classes in both datasets, we are able to apply the pre-trained model on our images without retraining it. For the SSD we use the output human bounding boxes and for the DeepLab we use the ouput human masks.

We treat single person keypoint detection as a semantic segmentation problem by generating the ground truth masks where pixels in a small region near the keypoints are set as the corresponding keypoint classes and others are set as background class. Then we trained a DeepLab v2\cite{deeplab_v2} model to learn this semantic segmentation representation.

During inference, we crop the human bounding box generated by the detector and adopt the DeepLab model to generate a pixel-wise saliency map of keypoints. If there are more than one region of the same keypoint type in the saliency map, we only take the one with the largest region area and eliminate the rest. Finally we get the final detection result by letting the centroid of keypoint regions in the saliency map be the final keypoints detection result.

We conducted the experiments by training the baseline model and all the training images we use are in the HKD training set. The quantitative results on the HKD validation set are in Tab.\ref{keypoint_baseline}

\begin{table}\small
\centering
\begin{tabular}{ccc}
\hline
Algorithms & mAP-12 & mAP-14 \\
\hline
Baseline(bbox) & 0.228 & 0.234 \\
Baseline(mask) & 0.226 & 0.233 \\
OpenPose\cite{openpose} & 0.296 & - \\
\hline
\end{tabular}
\caption{\small{The mAP score on the HKD validation set. mAP-12 score is evaluated on the 12 keypoints identical to MSCOCO and mAP-14 score is evaluated on all 14 keypoints in the HKD dataset}}\label{keypoint_baseline}
\end{table}

As we can see, OpenPose\cite{openpose}, the winner of MSCOCO 2016 keypoint competition\cite{2014MSCOCO}, scores only 0.296 mAP-12 on the HKD dataset. In the mean time, we provided a basic approach which adopts a top-down pipeline and scores a 0.228 mAP-12 value and a 0.234 mAP-14 value.

\section{Attribute based Zero-shot Recognition} 
\subsection{Overview}
Human beings can learn a new concept from descriptions without seeing it. Zero-shot recognition, which aims to recognize objects from novel unseen classes, is a promising approach to realize large-scale object recognition.
Significant progress\cite{xian2017zero,fu2017recent} has been achieved in zero-shot recognition. In most practices, ZSR is implemented by transferring knowledge from seen to unseen classes via auxiliary knowledge, e.g. attributes\cite{lampert2014attribute}, word vectors\cite{mikolov2013distributed} and gaze embeddings\cite{karessli2017gaze}. Compared to other types of auxiliary knowledge, attributes have good discrimination and interpretability. Many state-of-the-art ZSR results\cite{kodirov2017semantic,ye2017zero,zhang2016learning,zhang2016SPZSL,xian2016latent} have been achieved based on attributes.

\begin{table*}[]
\centering
\small
\begin{tabular}{l|l|l|l|l|l|l}
\hline
                 & LAD                 & CUB       & SUN       & aP/aY     & AwA       & ImageNet\_A            \\ \hline
Images           & 81,658              & 11,788    & 14,340    & 15,339    & 30,475    & 384,000*             \\ \hline
Classes          & 240                 & 200       & 717       & 32        & 50        & 384                 \\ \hline
Bounding Box     & Yes                 & Yes       & No        & Yes       & No        & Yes                \\ \hline
Attributes       & 359                 & 312       & 102       & 64        & 85        & 25                  \\ \hline
Annotation Level & 20 ins./class  & instance  & instance  & instance  & class     & 25 ins./class \\ \hline
\end{tabular}
\caption{\small{Statistics and comparison of different datasets. * means the estimated number.}}
\label{tab:statistics}
\end{table*}

However, there exists only a small number of image datasets annotated with attributes. The frequently used ones include Caltech-UCSD Birds-200-2011 (CUB)\cite{WahCUB_200_2011}, SUN Attributes (SUN)\cite{xiao2010sun}, aPascal/aYahoo (aP/aY)\cite{farhadi2009describing}, Animals with Attributes (AwA)\cite{farhadi2009describing} and ImageNet\_A\footnote{The authors provide attributes for 384 popular synsets in ImageNet. In this section, we use "ImageNet\_A" to refer this subset of ImageNet.}\cite{russakovsky2010attribute}  (see in Tab.\ref{tab:statistics}). Existing attribute datasets have three major limitations:
1) Small image numbers. The sum of images in CUB, aP\&aY, SUN and AwA datasets is only 72k. This is a small number compared to many object recognition datasets, e.g. the ImageNet\cite{NIPS2012_4824}, MSCOCO\cite{2014MSCOCO} and LSUN\cite{yu2015lsun}.
2) Lack of semantic attributes. Only low-level visual attributes (e.g. color, size, shape, texture) are annotated in CUB and ImageNet\_A.
3) Close to ImageNet. The categories in some datasets, e.g. AwA and aP\&aY, have a large overlap with ImageNet.
4) Serious distribution bias. For instance, 30\% classes in AwA have more than 10\% images in which the object is along with "person" . Such distribution bias may cause the inaccurate learning of some objects.
These limitations block the evaluation and improvement of ZSR methods.

\begin{figure}
  \centering
  \includegraphics[width=0.48\textwidth]{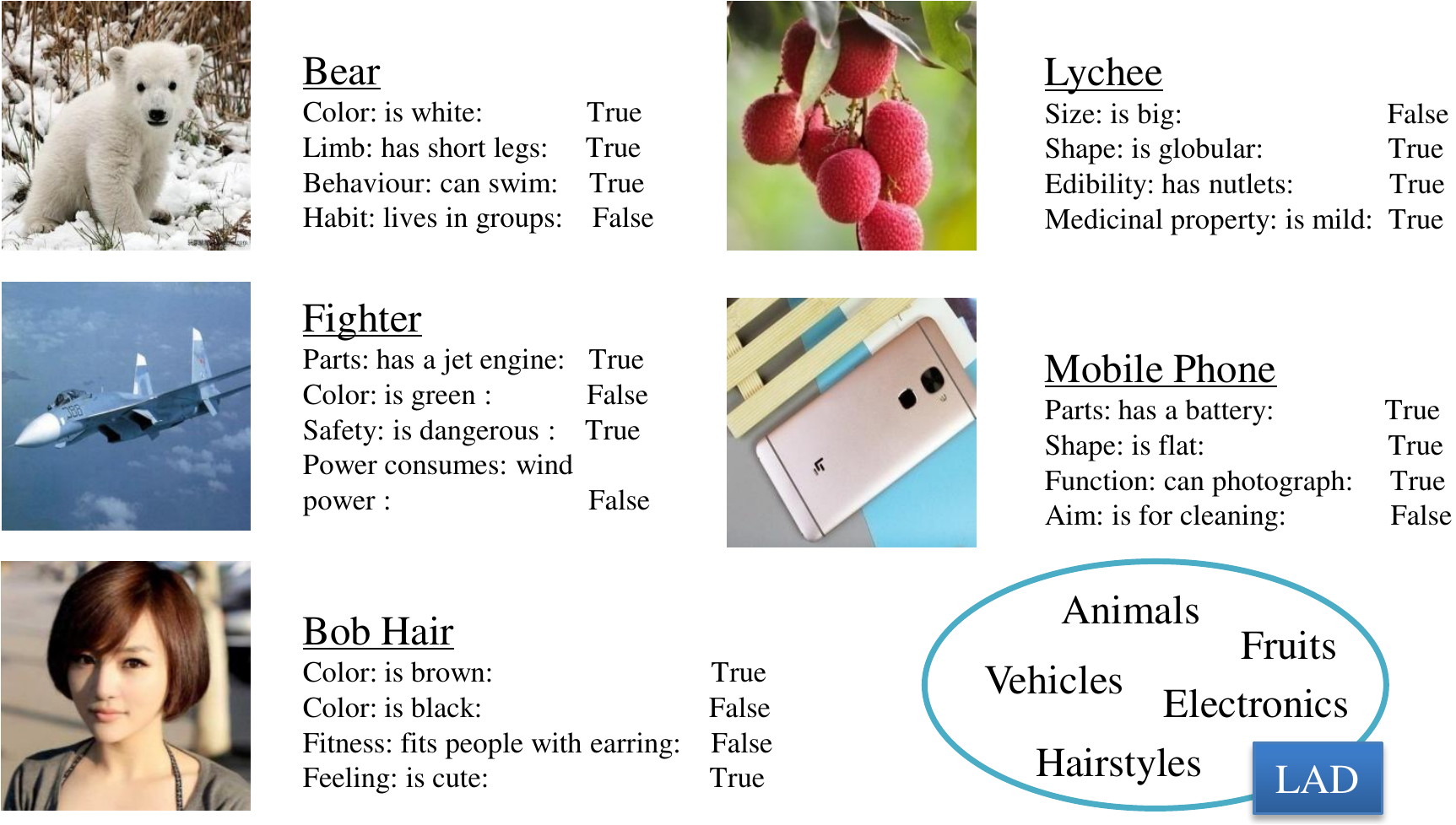}
  \caption{\small{Examples in our dataset. We annotate both visual attributes (the upper two) and semantic attributes (the bottom two).}}
  \label{fig:firstfig}
\end{figure}

We present a Large-scale Attribute Dataset (LAD) with rich semantic attributes (shown in Fig.\ref{fig:firstfig}) to promote the development of zero-shot recognition and other attribute-based tasks \cite{vaquero2009attribute,ma2000edgeflow,yao2016boosting,wu2017image, yu2010attribute}. Our dataset contains 81,658 images, 240 classes and 359 attributes.  Beyond low-level visual attributes
(e.g. colors, sizes, shapes), we also provide many semantic attributes. For example, we annotate attributes of diets and habits for animals, edibility and medicinal property for fruits, safety and usage scenarios for vehicles, functions and usage mode for electronics.

\subsection{Data Annotation \& Statistics}

To construct attribute based zero-shot recognition dataset, we first define the label list of all classes. Specifically, our dataset includes 240 classes. These classes can be divided into 5 subsets, namely animals, fruits, vehicles, electronics and hairstyles. The first four coarse-grained subsets contain 50 classes respectively, while the last fine-grained hairstyle subset contains 40 classes.

We crawl images for each class based on the search of the label and synonyms. Then, we filter these images and keep those images with only one foreground object matching the label. We also annotate the bounding box for every foreground object.

As our dataset includes 240 classes, it is unsuitable to design a list of many attributes for all classes. Hence, we design the attribute list for each subset. Specifically, we design 123, 58, 81, 75 and 22 attributes (359 in total) for animals, fruits, vehicles, electronics and hairstyles respectively.  Beyond low-level visual attributes (e.g. colors, shapes, sizes), we provide many semantic attributes (e.g. habits of animals, functions of electronics, feelings about hairstyles). These semantic attributes are human-concerned ones, however, not well investigated in previous vision tasks.

\begin{figure*}
\centering
\subfigure[Statistics of image numbers per class.]{
\label{fig:statistics.sub.1}
\includegraphics[width=0.36\textwidth]{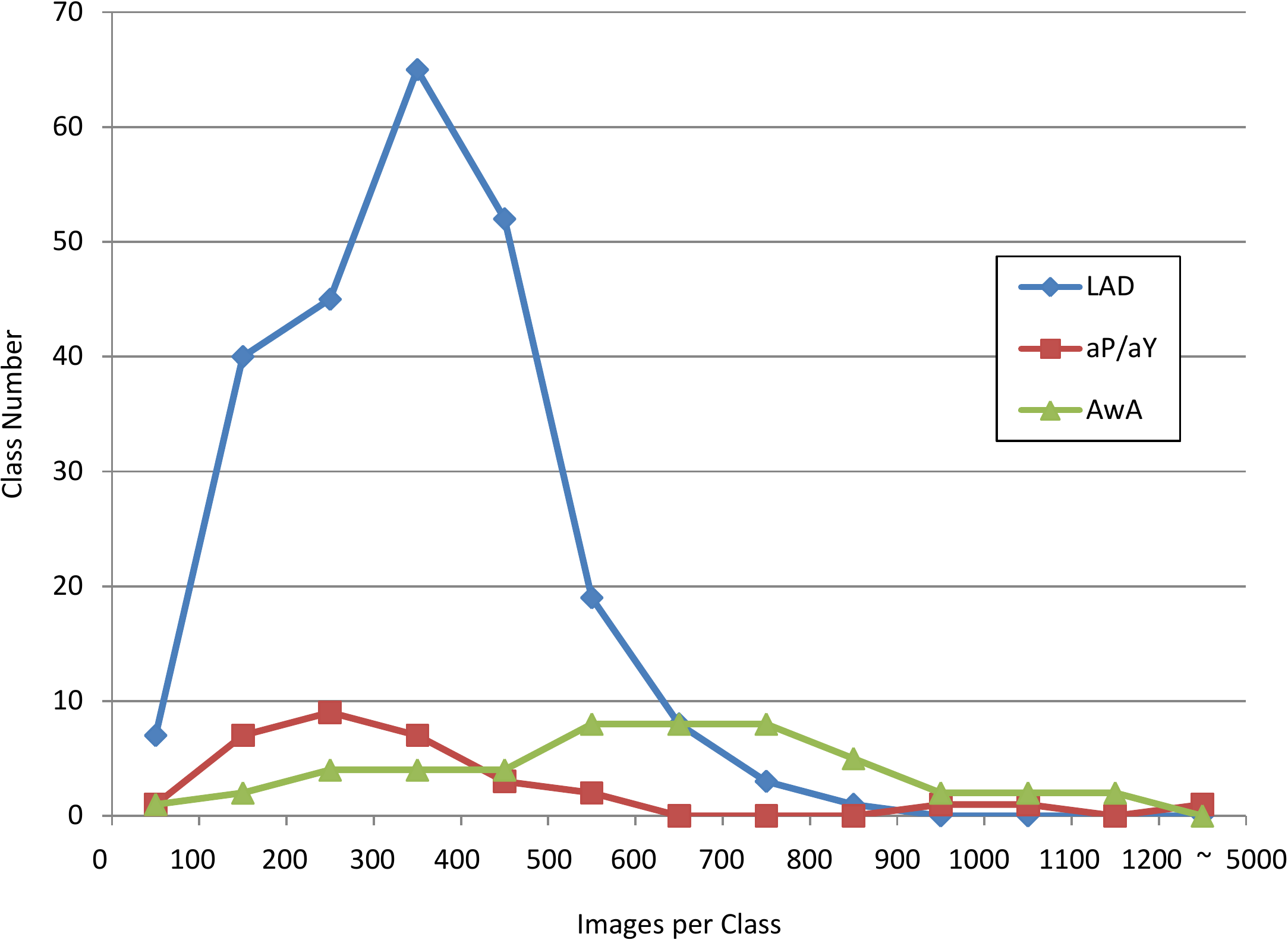}}
\subfigure[Statistics of class and attribute numbers.]{
\label{fig:statistics.sub.2}
\includegraphics[width=0.36\textwidth]{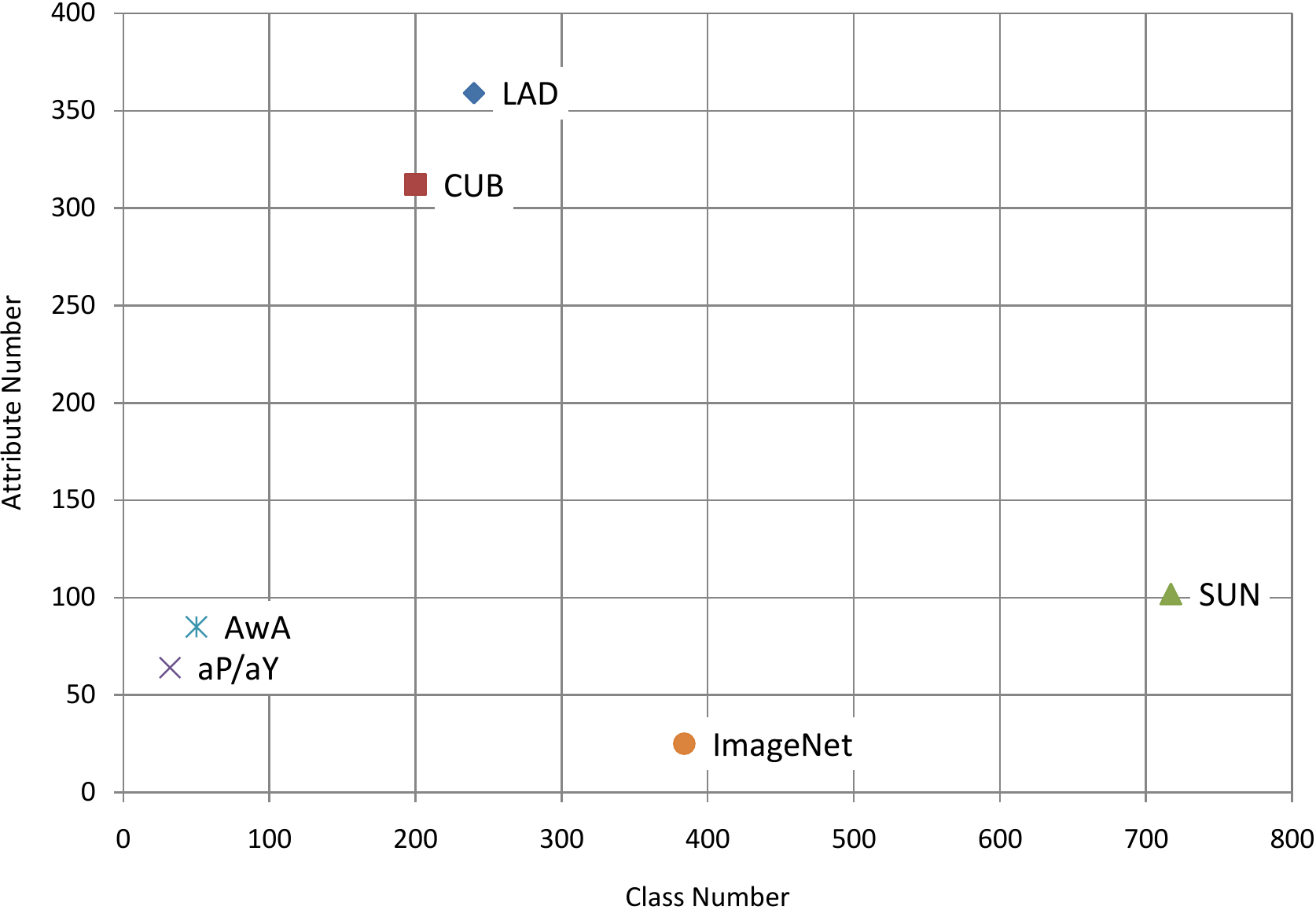}}
\caption{\small{Statistics of image, class and attribute numbers.}}
\label{fig:statistics}
\end{figure*}

Tab.\ref{tab:statistics} shows the statistics of image and annotation numbers of our dataset and others. Clearly, our dataset has the largest number of attributes. Our dataset has 81,658 images which is greater than the sum of CUB, SUN, aP/aY and AwA.
Fig.\ref{fig:statistics} illustrates the distribution of image numbers per class. Most classes in our dataset have around 350 images, which is greater than aP/aY dataset (around 250 images).

\subsection{Data Split}
We present a set of splits of seen/unseen classes for zero-shot recognition. We follow the traditional 80\%/20\% split ratio of seen/unseen classes. We shuffle these classes in each subset. Then we divide all 240 classes into 5 folders.
Every 20\% of these classes are chosen to be unseen classes and the rest are seen classes. In this way, we can obtain 5 random splits. For each subset, the ratio of seen/unseen is the same. We advocate to evaluate methods on the 5 splits and provide the mean accuracy.

For supervised learning of attributes and labels, we provide the split of training/testing data. We randomly select 70\% data from each class as training data and the rest 30\% are testing data. The validation data can be extracted from training data in experiments.

\subsection{Experiments}
\textbf{Baseline Methods.}
We implement zero-shot recognition experiments on our dataset using three basic methods, namely, SOC\cite{palatucci2009zero}, ESZSL\cite{romera2015embarrassingly} and MDP\cite{zhao2017zero}, which belongs to three popular frameworks. First, images
and labels are embedded into the image feature space (using ResNet pre-trained on ImageNet) and the semantic embedding spaces (using annotated attributes).
SOC tries to learn a linear mapping function from the image feature space to the semantic embedding space using seen class data. Then unseen instances are mapped to the semantic embedding space using the learned mapping function. These unseen instances are classified based on distances to the ground-truth unseen semantic embeddings using nearest-neighbour classifier. ESZSL learns a mapping that measures the compatibility between image features and semantic embeddings. MDP aims to learn the local structure of semantic embeddings. Then the structure is transferred to image feature space for synthesizing unseen image data. Labels of testing unseen images are predicted according to the distance to these synthesized data.

\begin{table}[]
\centering
\small
\begin{tabular}{l|l|l|l}
\hline
         & SOC      & ESZSL & MDP      \\ \hline
1        & 31.05    & 42.17 & 46.96       \\ \hline
2        & 31.27    & 46.82 & 46.83         \\ \hline
3        & 34.64    & 42.49 & 51.41         \\ \hline
4        & 34.21    & 41.96 & 48.61        \\ \hline
5        & 36.57    & 43.72 & 49.08         \\ \hline
Ave      & 33.55    & 43.43 & {48.58}     \\ \hline
\end{tabular}
\caption{\small{Comparison of zero-shot recognition methods on our dataset.}}
\label{tab:zsrcomp}
\end{table}

\textbf{Experimental Methods.}
Experimental results are shown in Tab.\ref{tab:zsrcomp}. We can find that the zero-shot recognition accuracies on the five splits are balanced. MDP achieves the best performance, averagely 48.58\%. The runner-up method is ESZSL, whose average recognition accuracy is 43.43\%. The average recognition accuracy of SOC is 33.55\% which is around 15\% lower than MDP.

\section{Image Captioning for Chinese}

\subsection{Overview}
Image captioning has long been a challenging problem in computer vision and natural language processing. A great image model must capture not only primary objects contained in an image, but also the relationship between objects, their attributes, or the activities they are involved in. Moreover, the image captioning task requires that these semantic knowledge to be organized and conveyed in textual description, and therefore a language model is also needed.

Early approaches to tackle this issue could be roughly divided into two types: template-based methods\cite{2015_CVPR_visual_concepts,2013babytalk,2012collective,2011corpus} and retrieval-based approaches\cite{2015language_models,2010every}. The first approaches start from detecting object, action, scene and attributes in images and then combined them by language models. The second approaches retrieve the visually similarity images from a large database, and then transfer the captions of retrieved images to fit the query image.

Recently, the encoder-decoder framework\cite{2015_CVPR_FeifeiLi,2015_CVPR_show_and_tell,2016_CVPR_Caption_Semantic_Attention,2017_CVPR_sca_cnn} and the reinforcement learning framework\cite{2017_CVPR_Reinforcement} have been introduced to image captioning. Researchers adopted encoder-decoder framework because ¡°translating¡± an image to a sentence was analogous to the task in machine translation. Approaches following this framework generally encode an image as a single feature vector by convolutional neural networks, and then feed such vector into recurrent neural networks to generate captions. Reinforcement learning framework is based on decision-making which utilizes a "policy network" and a "value network" to collaboratively generate captions.

\begin{figure}[htp]
\begin{center}
   \includegraphics[width=1\linewidth]{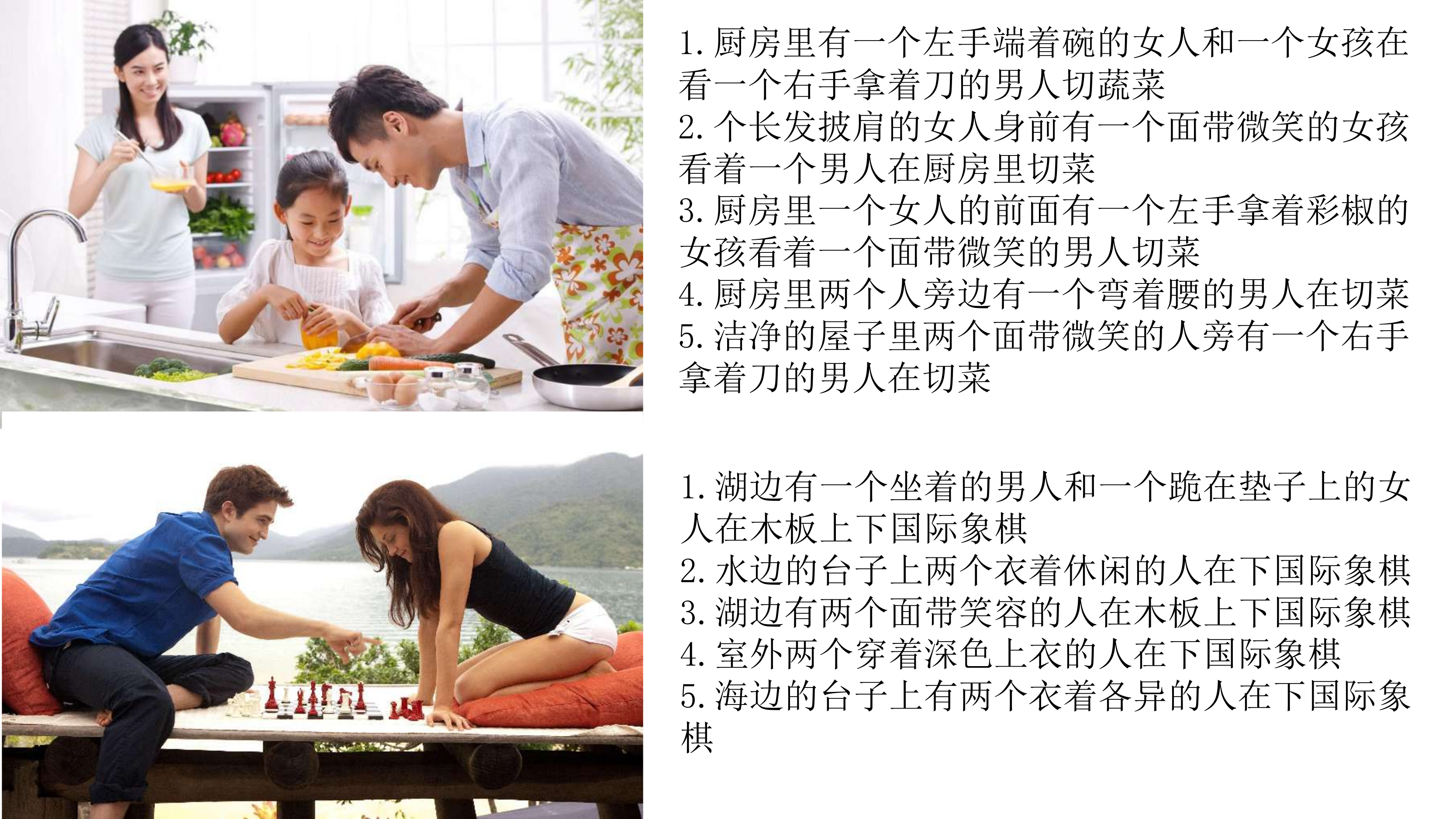}
\end{center}
   \caption{\small{Examples of ICC training dataset.}}
\label{caption_train}
\end{figure}

Although much of the progress have been made possible by the availability of image caption datasets such as Pascal VOC 2008\cite{2010every}, Flickr8k\cite{2015_flickr8k}, Flickr30k\cite{2014Flickr30k}, MSCOCO\cite{2014MSCOCO} and SBU\cite{2011im2text} datasets, captions in existing datasets were all labeled in English. These datasets contain 8,000, 31,000 and 300,000 images respectively and each is annotated with 5 English sentences. To promote progress in this area, we created the image Chinese captioning (ICC) dataset (see in Fig.\ref{caption_train}). To our knowledge, the ICC dataset is the largest image captioning dataset whose sentences are labeled in Chinese.

The rest of this section is organized as follows. Firstly, we describe the process of collecting Chinese captions for the ICC dataset. Secondly, we analyze the properties of the ICC dataset. Thirdly, we introduce a baseline for the ICC dataset. Finally, we perform experiments to assess the effectiveness of the baseline model using several metrics.

\subsection{Dataset Statistics}
We analyze the properties of the ICC dataset in comparison to several other popular datasets. The statistics of the datasets are shown in Tab.\ref{caption_datasets}.

\begin{figure*}
  \centering
  \includegraphics[width=0.9\linewidth]{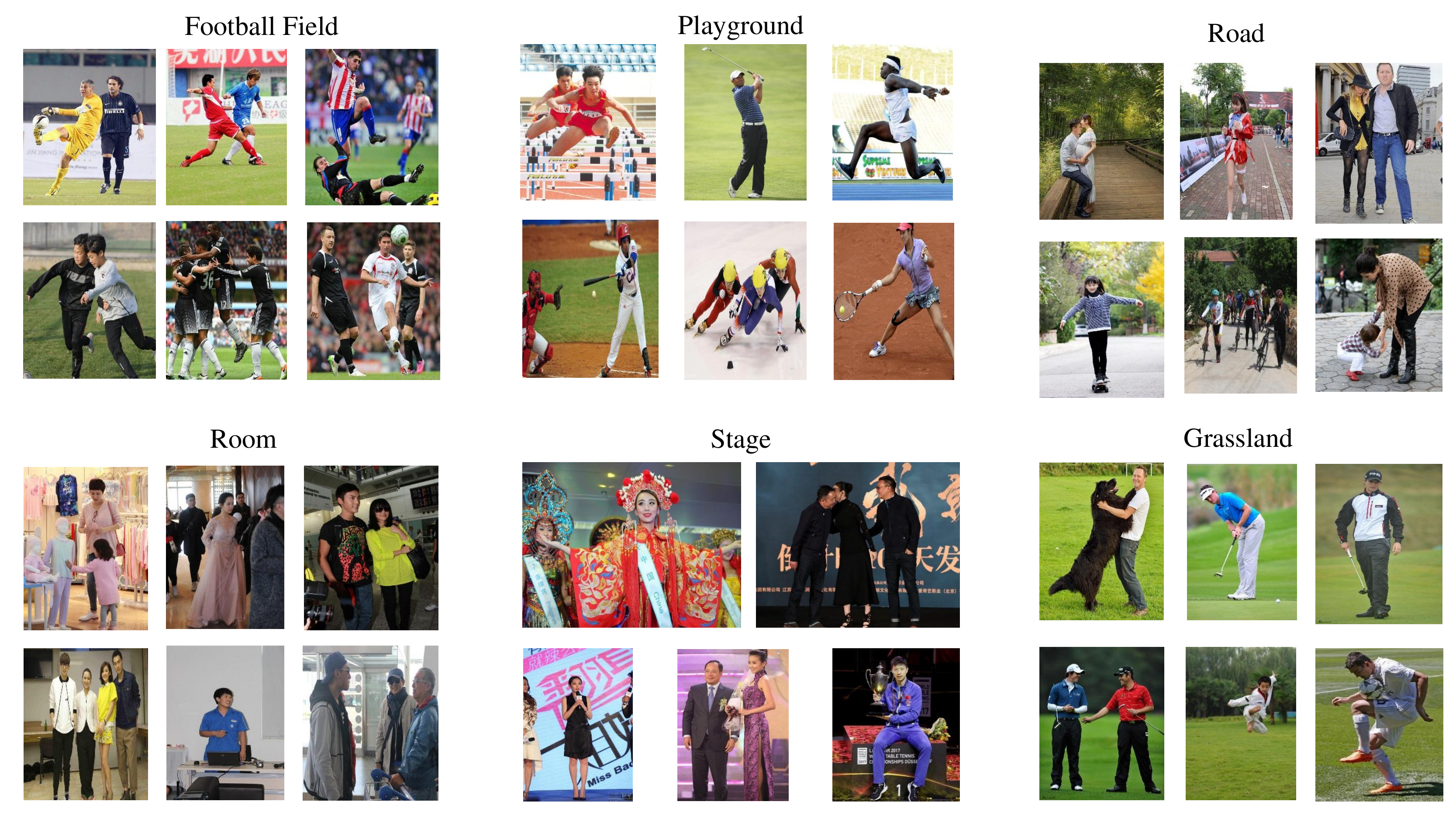}
  \caption{\small{Examples of different scenes.}}
  \label{scenes_places}
\end{figure*}

\subsection{Data Annotation}
The pipeline to gather data for the ICC dataset can be divided into two major parts, which are image selection (similar as HKD) and Chinese caption labeling.

Apart from the HKD image selection principles, we add two more rules. Firstly, the image should be easy to be described using only one sentence. Secondly, the image should contain multiple objects in complex scenes. For example, an image contains only one person standing with no other significant poses is less likely to be selected than the one with two people hugging.

The ICC dataset contains five reference captions for every image, which were labelled by 5 different native speakers in China using Chinese language. Each of our captions is generated using human subjects which are similar to the ones in\cite{2015mscoco_caption}.

There are three principles that guide image caption annotation. Firstly, the annotations should include but not limited to key objects/attributes, locations and human actions. Secondly, the sentences should be fluent. Thirdly, the use of Chinese idioms or descriptive adjectives is encouraged.

\begin{table}[htp]\small
\centering
\begin{tabular}{l|c|c|c|c|c}
\hline
dataset name & train & valid & test-1 & test-2 & language\\
\hline
Pascal VOC\cite{2010every} & - & - & 1K & - & English\\
Flickr8k\cite{2015_flickr8k} & 6K & 1K & 1K & - & English\\
Flickr30k\cite{2014Flickr30k}& 28K & 1K & 1K & - & English\\
MSCOCO\cite{2014MSCOCO} & 82K & 40K & 40K & - & English\\
SBU\cite{2011im2text} & 1M & - & - & - & English\\
ICC & 210K & 30K & 30K & 30K & Chinese\\
\hline
\end{tabular}
\caption{\small{Statistics and comparison of different datasets.}}\label{caption_datasets}
\end{table}

The number of captions is 1,050,000 captions for 210,000 images in training, 150,000 captions for 30,000 images in validation, 150,000 captions for 30,000 images in testing-1 and 150,000 captions for 30,000 images in testing-2. ICC is the largest dataset whose captions are in Chinese and it is the first to provide two different test datasets which can better evaluate if the algorithm is overfitting.

In ICC training data there are more than 200 scenes and places such as "football field" and "grassland" (see in Fig.\ref{scenes_places}), 150 actions such as "sing" and "run". ICC dataset contains most of common daily scenes in which a person usually appear.

\begin{table*}\small
\centering
\begin{tabular}{l|c|c|c|c|c|c|c}
\hline
Algorithm & BLEU-1 & BLEU-2 & BLEU-3 & BLEU-4 & CIDEr & METEOR & ROUGE\\
\hline
Baseline & 0.765 & 0.648 & 0.547 & 0.461 & 1.425 & 0.370 & 0.633\\
\hline
\end{tabular}
\caption{Scores of caption baseline for ICC testing-1.}\label{caption_baseline_results}
\end{table*}

\subsection{Baseline Model}
We adapt "show and tell", a popular encoder-decoder model for image captioning\cite{2015_CVPR_show_and_tell}, as our base model. One difference is that we use the "Jieba" Chinese word segmentation module during preprocessing, instead of the English tokenization module used in "show and tell".

This model directly maximize the probability of the correct description given the image by using the following formulation:
$$\theta^* = \arg \max_\theta \sum_{(I,S)}\log{p(S|I;\theta)}$$
where $\theta$ are the parameters of our model, $I$ is an image, and $S$ is a correct transcription.

\subsection{Experimental Results}
To quantitatively evaluate how well the base model learns to generate Chinese captions, experiments were conducted on the ICC testing-1 dataset which contains 30,000 images. All the reported results are computed on the metrics BLEU\cite{2002bleu}, METEOR\cite{2014meteor}, ROUGE\cite{2004rouge} and CIDEr\cite{2015cider} respectively, which are commonly used together for fair and thorough performance measurement.

In Tab.\ref{caption_baseline_results}, we provide a result summary of our baseline model. We achieve reasonable performance on ICC in most evaluation metrics.

\begin{figure}
  \centering
  \includegraphics[width=1\linewidth]{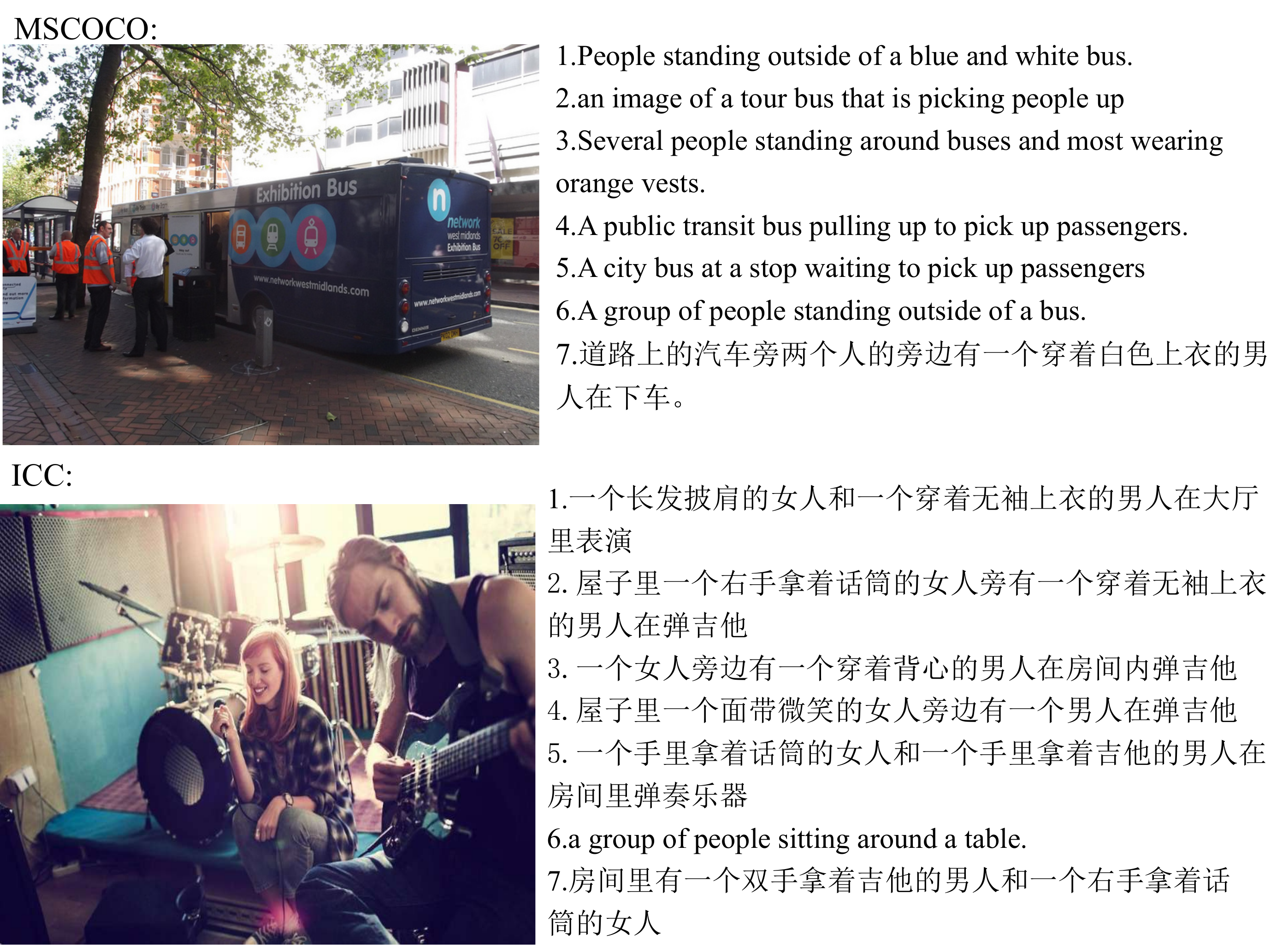}
  \caption{\small{Results of baseline model for ICC and MSCOCO.}}
  \label{caption_result}
\end{figure}

In Fig.\ref{caption_result}, captions of MSCOCO and ICC datasets are shown respectively. Both of the first 5 captions are written by human. The sixth caption is generated by the baseline model trained on MSCOCO dataset and the seventh caption is generated by the same model trained on ICC dataset. The same model trained on ICC dataset produces better performance than the one trained on MSCOCO dataset in most cases. For example, the seventh caption in the first image, which translates as "Beside two people next to a car on the road, there is a man wearing a white shirt getting off the car", and the seventh caption in the second image, which "In the room there are a man holding a guitar with two hands and a woman holding a microphone with her right hand", both provide much more descriptive details than the captions generated by model trained on MSCOCO. The results show that the captions in ICC dataset could provide more context information.

\section{Conclusion}
In this paper, we propose a new dataset with rich annotations, for training and evaluating methods. Utilizing over 185,000 worker hours, a vast collection of images was collected, annotated and organized to provide three new, large volume datasets for human keypoint detection, attribute based zero-shot recognition and image Chinese captioning.  Among these three datasets, intersection is significant, and people can cross reference low level image annotations such as class labels to high level segment annotations such as captioning. This provides a good benchmark for evaluating and improving methods for these three tasks and other possible tasks to cross correlate different levels of information. On our dataset, we also provide basic statistical tests and base line models to prove the basic validity and first insight.

There are several promising directions for future annotations on our dataset. For example, currently the human keypoint dataset only includes skeletal keypoints of human figures, but annotating "expression" or "action" may provide more information that can be useful for even higher-level visual tasks, such as pose estimation. Moreover, we currently only collect images containing human beings for image Chinese captioning dataset, but collecting other classes may provide better relationship between objects.

To download and learn more about AIC dataset, please refer to the project website\footnote{\url{https://challenger.ai/}}. Some code is released online\footnote{\url{https://github.com/AIChallenger/AI_Challenger}}.

{\small
\bibliographystyle{ieee}
\bibliography{egbib}
}

\end{document}